\documentclass{bmvc2k}

\usepackage{multirow}
\usepackage{booktabs}
\usepackage{colortbl}
\definecolor{mygray}{gray}{0.95}

\title{MVT: Multi-view Vision Transformer for 3D Object Recognition}

\addauthor{Shuo Chen \\
Tan Yu \\
Ping Li}{{shanshuo1992, tanyuuynat, pingli98}@gmail.com}{1}

\newcommand\blfootnote[1]{%
  \begingroup
  \renewcommand\thefootnote{}\footnote{#1}%
  \addtocounter{footnote}{-1}%
  \endgroup
}

\addinstitution{
Cognitive Computing Lab, \\
Baidu Research
}

\runninghead{Chen, Yu, Li}{Multi-view Vision Transformer}

\usepackage{multirow}
\usepackage{graphicx}
\usepackage{wrapfig}
\usepackage{booktabs,amsfonts,dcolumn}
\usepackage{colortbl}
\definecolor{mygray}{gray}{0.95}

\begin{document}

\maketitle
\blfootnote{\hspace{-0.55cm} The work of Shuo Chen was conducted as an intern at Baidu. He is a PhD student at the University of Amsterdam.}

\begin{abstract}
Inspired by the great success achieved by CNN in image recognition, view-based methods applied CNNs to model the projected views for 3D object understanding and achieved excellent performance. Nevertheless, multi-view CNN models cannot model the communications between patches from different views, limiting its effectiveness in 3D object recognition. Inspired by the recent success gained by vision Transformer in image recognition, we propose a Multi-view Vision Transformer (MVT) for 3D object recognition. Since each patch feature in a Transformer block has a global reception field, it naturally achieves communications between patches from different views. Meanwhile, it takes much less inductive bias compared with its CNN counterparts. Considering both effectiveness and efficiency, we develop a global-local structure for our MVT. Our experiments on two public benchmarks, ModelNet40 and ModelNet10, demonstrate the competitive performance of our MVT.
\end{abstract}

\section{Introduction}
In the past decade, we have witnessed the great success achieved by convolutional neural network~\cite{krizhevsky2012imagenet,he2016deep} in image understanding. Inspired by its success in understanding 2D images, several works attempt to deploy CNN in 3D object understanding, achieving excellent performance. 
These methods can be coarsely divided into three groups: view-based methods~\cite{su2015multi,wang2017dominant,feng2018gvcnn,kanezaki2018rotationnet,DBLP:journals/tip/HanSLVLZHC19,DBLP:journals/tip/HanLLVLZHC19}, volume-based methods~\cite{wu20153d,maturana2015voxnet,qi2016volumetric,meng2019vv}, and point-based methods~\cite{pointnet,pointnet++,3Dmfv}. Among them, view-based methods are closely related to 2D image understanding. View-based methods project a 3D object into multiple views and model each view through the model original used in modelling 2D images.  Benefited from pre-training on the large-scale 2D image dataset such as ImageNet~\cite{deng2009imagenet}, they achieve competitive performance compared with their volume-based and point-based counterparts.

MVCNN~\cite{su2015multi} is the pioneering view-based method. It extracts each view features through a vanilla 2D CNN and aggregates the view features through sum-pooling.  The following works~\cite{wang2017dominant,feng2018gvcnn,kanezaki2018rotationnet,DBLP:journals/tip/HanSLVLZHC19,DBLP:journals/tip/HanLLVLZHC19,Wei_2020_CVPR} seek to find a more effective way to aggregate the view features. Specifically, RCPCNN~\cite{wang2017dominant} and GVCNN~\cite{feng2018gvcnn} group views into multiple sets and conduct pooling within each set. Seqviews2seqlabels~\cite{DBLP:journals/tip/HanSLVLZHC19} and 3D2SeqViews~\cite{DBLP:journals/tip/HanLLVLZHC19} model the view order through recurrent neural network. View-GCN~\cite{Wei_2020_CVPR} models the view-based relations through graph convolution network. MHBN~\cite{yu2018multi} and MVLADN~\cite{DBLP:journals/tip/YuMYY21} observe the limitations of view-based pooling, formulate the view-based $3$D object recognition into a set-to-set matching problem, and investigate in patch-level pooling. Nevertheless,  view-based pooling and patch-based pooling methods only fuse the visual features from different views in the last pooling layer. There are no interactions between visual features from different views in previous layers. This configuration leads to the fact that a  patch can only have a local perception field and fails to perceive patches in other views. Relation Network~\cite{mvrelations} enhances each patch feature by patches from all views, achieving better performance than the above-mentioned view-based pooling and patch-based pooling methods.

\begin{figure}
     \centering
    \includegraphics[width=.75\linewidth]{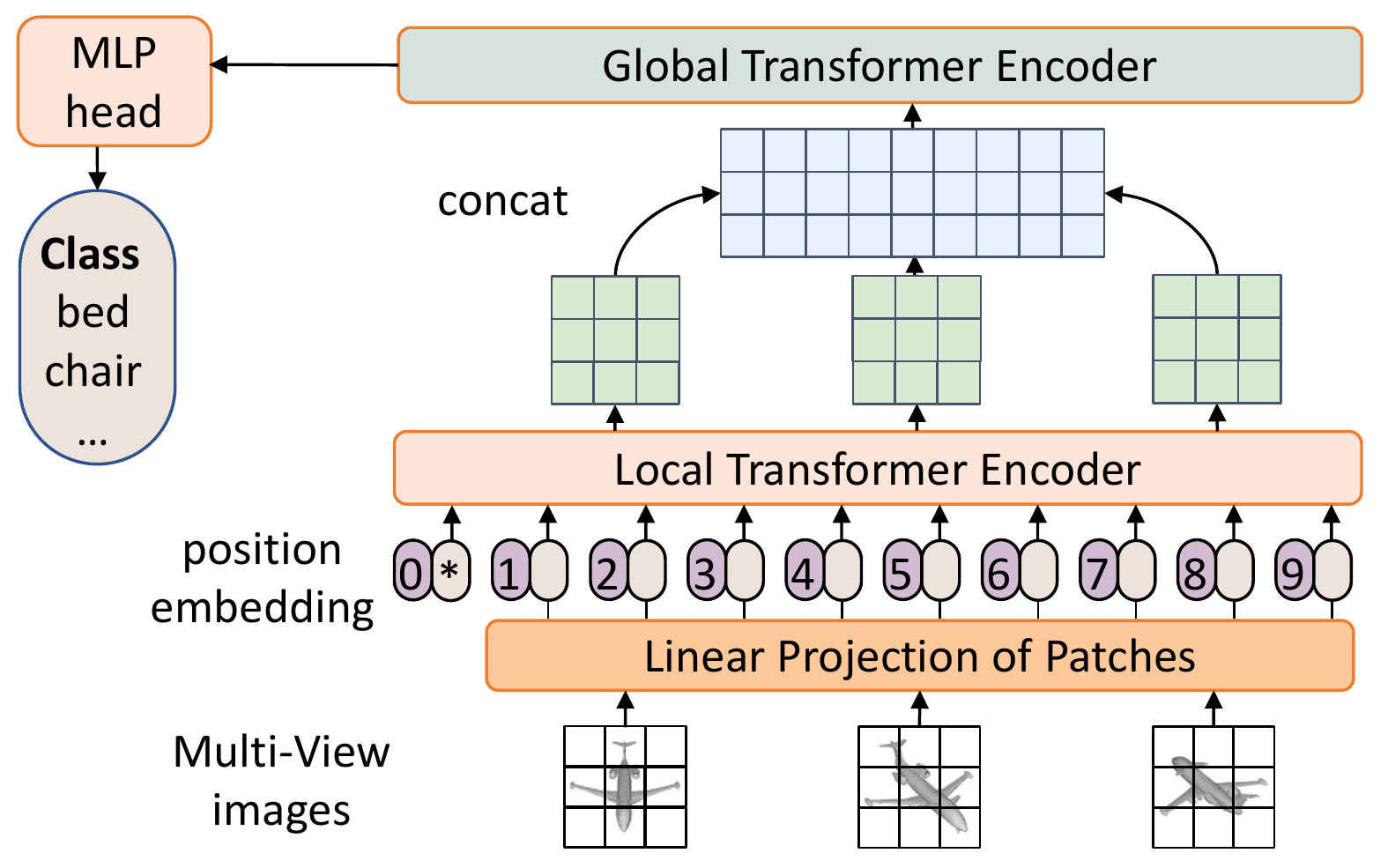}
     \caption{
    The architecture of the proposed Multi-view Vision Transformer (MVT). Each view is split into non-overlap patches. Each patch is projected into an embedding vector,  which is added to a position embedding.   Patch embedding vectors from each view are fed into the local Transformer encoder for communications between patches within the view. Then the outputs of the local Transformer for all views are concatenated into a global set, which is fed into the global Transformer encoder for communications between patches from different views.  After that, the set of attended patch features from the output of the global Transformer encoder are sum-pooled into a global representation for the $3$D object.  We finally use the MLP head for classification.
     }
     \label{fig:overview}
\end{figure}

In this work, inspired by the great success achieved by vision Transformer~\cite{dosovitskiy2020image,touvron2020training}, we propose a multi-view vision Transformer (MVT) to empower each patch to have the global reception field to perceive the visual content of all views from a 3D object. It adopts a pure-Transformer architecture and thus takes much less inductive bias compared with its CNN counterparts~\cite{dosovitskiy2020image}. Considering the total number of patches is largely due to multiple projected views, simply concatenating all patches will generate an extremely long sequence, leading to an expensive computational cost.  Taking both effectiveness and efficiency into consideration, we devise a local-global structure, as visualized in Figure~\ref{fig:overview}. In the local Transformer encoder, we adopt Transformer to process the patches within each view individually. In the global Transformer encoder, we merge patch features from all views and feed them together into Transformer layers for the global reception field. Using such a simple and elegant architecture, we achieve state-of-the-art recognition accuracy on public benchmarks, including ModelNet40 and  ModelNet10.

\section{Related Works}
\textbf{3D object recognition}. Existing mainstream 3D object recognition methods can be categorized into three groups:  volume-based methods~\cite{wu20153d,maturana2015voxnet,qi2016volumetric,meng2019vv}, point-based methods~\cite{pointnet,pointnet++,3Dmfv} and view-based methods~\cite{wang2017dominant,feng2018gvcnn,kanezaki2018rotationnet,DBLP:journals/tip/HanSLVLZHC19,DBLP:journals/tip/HanLLVLZHC19,Wei_2020_CVPR},. Among them, volume-based methods quantize the 3D object into regular voxels, and conduct  3D convolutions on voxels. Nevertheless, 3D convolution is computationally expensive when the resolution is high. For satisfactory efficiency,  volume-based methods  normally conduct low-resolution quantization, inevitably leading to information loss.
In parallel, point-based methods directly model the cloud of points, efficiently achieving competitive performance. View-based methods project a 3D object into multiple 2D views. They model each view through the vision backbone for image understanding to obtain view features or patch features. Our work can be categorized into view-based methods. Thus, we mainly review view-based methods here. 

MVCNN~\cite{su2015multi} is one of the earliest works exploiting convolutional neural network (CNN) for modelling multiple views. It aggregates the view features from CNN through max-pooling.  MVCNN-MultiRes~\cite{qi2016volumetric} exploits views projected from multi-resolution settings, boosting the recognition accuracy. Pairwise~\cite{johns2016pairwise} decomposes the sequence of projected views into several pairs and models the pairs through CNN.  GIFT~\cite{bai2016gift} represents each 3D object by a set of view features and determines the similarity between two 3D objects by matching two sets of view features through a devised matching kernel. RotationNet~\cite{kanezaki2018rotationnet} considers the viewpoint of each project and treats viewpoints as latent variables to boost the recognition performance. 
RCPCNN~\cite{wang2017dominant} groups views into multiple sets and concatenate the set features as the 3D object representation. GVCNN~\cite{feng2018gvcnn} also groups views into multiple sets. It adaptively assigns a higher weight to the group containing crucial visual content to suppress the noise. Seqviews2seqlabels~\cite{DBLP:journals/tip/HanSLVLZHC19} and 3D2SeqViews~\cite{DBLP:journals/tip/HanLLVLZHC19} exploit the view order besides visual content through recurrent neural network.   View-GCN~\cite{Wei_2020_CVPR} models the relations between views by a graph convolution network. MHBN~\cite{yu2018multi} and MVLADN~\cite{DBLP:journals/tip/YuMYY21} investigate in pooling patch-level features to generate the 3D object recognition.  Relation Network~\cite{mvrelations} enhances each patch feature by patches from all views through a reinforcement block plugged in the rear of the network. Our method has a similar spirit but needs much less inductive bias and only takes a standard Transformer to achieve the communications between patches of different views.

\noindent \textbf{Vision Transformer.} Inspired by the great success achieved by Transformers~\cite{vaswani2017attention} in natural language processing,  vision Transformer~(ViT)~\cite{dosovitskiy2020image} is proposed. It crops an image into multiple non-overlap patches and feeds the cropped patches into a stack of Transformer layers.  Compared with CNN models, each patch in ViT has a global reception field. Meanwhile, ViT has much less image-specific inductive bias than CNN models.
By pre-training on huge-scale datasets, ViT has achieved comparable accuracy compared with its CNN counterparts.  DeiT~\cite{touvron2020training} proposes a data-efficient approach using an improved optimizer,  more advanced data augmentation, and training tricks. PVT~\cite{wang2021pyramid} and PiT~\cite{heo2021pit} bring back inductive bias in CNN and exploit the pyramid structure to shrink the spatial size progressively.   
T2T~\cite{yuan2021tokens} and TNT~\cite{han2021transformer} focus on improving the effectiveness of modeling local structure within patches.
Swin~\cite{liu2021swin} and Twin~\cite{chu2021twins} exploit the locality and sparsity to achieve a better trade-off between effectiveness and efficiency. CvT~\cite{wu2021cvt} and Container~\cite{gao2021container} exploit a hybrid structure combining Transformer and convolution. Unlike the methods mentioned above that exploit Transformer for 2D image understanding, we investigate its effectiveness in 3D object recognition.

\section{Preliminary}

\begin{wrapfigure}{r}{0.4\textwidth}
\vspace{-0.7in}
  \begin{center}
    \includegraphics[scale=0.8]{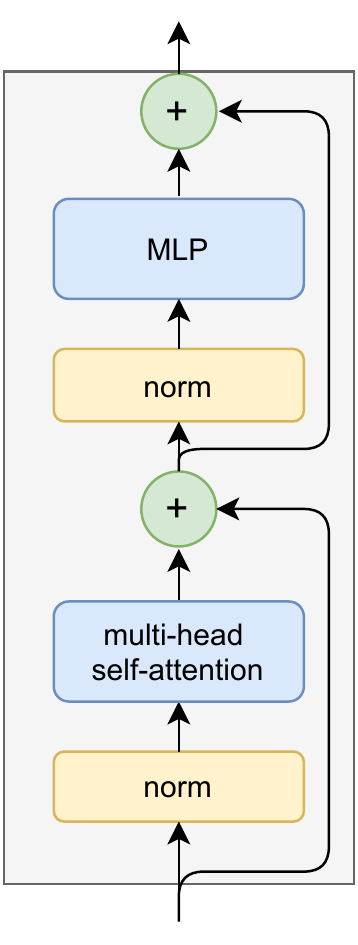}
  \end{center}
 \vspace{-0.25in}
\caption{The structure of a Transformer block. It consists of layer-normalization (norm) layers, a multi-layer perceptron (MLP) module and a multi-head self-attention module.}
\label{fig:trans}\vspace{-0.15in}
\end{wrapfigure}
In this section, we briefly introduce the structure of Transformer~\cite{vaswani2017attention} block as visualized in Figure~\ref{fig:trans}. It consists of two layer-normalization (LN) layers, a multi-head self-attention (MSA) module, and a multi-layer perceptron (MLP) module. 

\vspace{0.1in}
\noindent \textbf{Multi-head self-attention (MSA) module.} Let us denote the inputs by $\mathbf{X} \in \mathbb{R}^{N \times D}$ where $N$ is the number of input vectors and $D$ is the dimension of each input vector. MSA maps $\mathbf{X}$ into the queries $\mathbf{Q}^{N\times D_Q}$, keys $\mathbf{K} \in \mathbb{R}^{N\times D_K}$ and values $\mathbf{V}\in \mathbb{R}^{N\times D_V}$ through three fully-connected layers, where $D_Q = D_K = D_V$. Then it splits $\mathbf{Q}$, $\mathbf{K}$ and $\mathbf{V}$ into $M$ heads:
\begin{equation*}
    \mathbf{Q} \to [\mathbf{Q}_1,\cdots, \mathbf{Q}_M],~\mathbf{K} \to [\mathbf{K}_1,\cdots, \mathbf{K}_M],~\mathbf{V} \to [\mathbf{V}_1,\cdots, \mathbf{V}_M],
\end{equation*}
where, for $m = 1, \dots, M$, we have
\begin{equation*}
    \mathbf{Q}_m \in \mathbb{R}^{N\times \frac{D_Q}{M}},~\mathbf{K}_m \in \mathbb{R}^{N\times \frac{D_K}{M}},~\mathbf{V}_m \in \mathbb{R}^{N\times \frac{D_V}{M}}.
\end{equation*}
Then the self-attention operation is conducted on each query-key-value triplet  $\{\mathbf{Q}_m, \mathbf{K}_m, \mathbf{V}_m\}$ ($m = 1,\dots,M$) and generates the attended features:
\begin{equation*}
    \mathbf{Y}_m =  \mathrm{softmax}(\frac{\mathbf{Q}_m\mathbf{K}^{\top}_m}{\sqrt{D_k}}) \mathbf{V}_m,~ m = 1,\dots,M.
\end{equation*}
The attended features from each head are concatenated to obtain the final output of the self-attention module:
\begin{equation*}
    \mathbf{Y} \gets [\mathbf{Y}_1,\cdots,\mathbf{Y}_M] \in \mathbb{R}^{N \times D_V}.
\end{equation*}

\noindent \textbf{Layer Normalization (LN)}~\cite{ba2016layer} is widely used in Transformer-based architecture for training stability.  Given a $D$-dimension feature vector $\mathbf{x}  = [x_1,\cdots,x_D]$, it  computes the mean $\mu$ and the standard deviation $\epsilon$ by
\begin{equation*}
\mu = \frac{1}{D}\sum_{i=1}^D x_i,~\epsilon = \sqrt{\frac{1}{D}\sum_{i=1}^D (x_i-\mu)^2}.
\end{equation*}
Then a linear operation is conducted on each element of $\mathbf{x}$:
\begin{equation}
    \hat{x}_i = \gamma \frac{x_i-\mu}{\epsilon} + \beta,
\end{equation}
where $\beta$ and $\gamma$ are learnable parameters for affine transform.

\vspace{0.1in}

\noindent \textbf{Multi-layer perceptron (MLP).} The MLP is normally plugged after self-attention module to operate on each input separately~\cite{vaswani2017attention}. It consists of two fully-connected layers with a bottleneck structure and an activation layer for for non-linearity. Specifically, for each feature $\mathbf{x} \in \mathbb{R}^{D}$, MLP enhances $\mathbf{x}$ by 
\begin{equation}
    \mathrm{MLP}(\mathbf{x}) = \sigma(\mathbf{x}\mathbf{W}_1+\mathbf{b}_1)\mathbf{W}_2 + \mathbf{b}_2,
\end{equation}
where $\mathbf{W}_1 \in \mathbb{R}^{D \times rD}$ and $\mathbf{b}_1 \in \mathbb{R}^{rD}$ are weights of the first fully-connected layer,  which increase the feature dimension from $D$ to $rD$ and $r>1$ is termed as expansion ratio. Meanwhile, $\mathbf{W}_2 \in \mathbb{R}^{rD \times D}$ and $\mathbf{b}_2 \in \mathbb{R}^{D}$ are weights of the second fully-connected layer,  which decreases the feature dimension from $rD$ back to $D$.

\vspace{-0.1in}
\section{Method}
In this section, we introduce the proposed multi-view vision Transformer. We first clarify the input of the model and then illustrate the details of the model.

\vspace{-0.1in}

\subsection{Input} 

For each 3D object, we project it into $L$ views $\{V^1,\cdots, V^L\}$. Each view $V^j$ ($j  = 1, \dots, L$) is of  $W\times H \times 3$ size. 
For each view, we crop it into $w\times h$ non-overlap patches and each patch is of $p\times p \times 3$ size. Each patch is unfolded into a vector $\mathbf{p} \in \mathbb{R}^{3p^2}$. We denote  the $i$-th patch in the view $V^j$ by $\mathbf{p}_{i}^j$, where $i\in[1,wh]$ and $j\in[1,L]$. 
For each $\mathbf{p}_{i}^j$, we map it into a $D$-dimension vector through a fully-connected layer and obtain 
\begin{equation}
    \mathbf{x}_i^j \gets \mathbf{W}_0 \mathbf{p}_i^j,~i = 1,\dots, wh,~ j = 1, \dots, L, 
\end{equation}
where $\mathbf{W}_0 \in \mathbb{R}^{D\times 3p^2}$ is the weight matrix.  Meanwhile,  a position embedding $\mathbf{p}_i$ is learned for each spatial location $i\in[1,wh]$. Then the patch feature is obtained by summing up its visual feature and position embedding:
\begin{equation}
    \mathbf{z}_i^j \gets \mathbf{x}_i^j + \mathbf{p}_i.
\end{equation}
Note that, $ \mathbf{p}_i$ is only dependent on the spatial location ($i$) and is shared among different views.
Like BERT's \texttt{[class]} token~\cite{devlin2019bert}, for each view $V^j$, we additionally devise a special token, $\mathbf{z}_0^j$, which is a learnable embedding whose state at the output of the Transformer encoder serves as the view representation. We merge the special token $\mathbf{z}_0^j$ and  the patch features $\{\mathbf{z}_i^j\}_{i=1}^{wh}$ into a matrix $\mathbf{Z}^j$ defined as 
\begin{equation}
\label{input}
    \mathbf{Z}^j = [\mathbf{z}_0^j,\mathbf{z}_1^j,\cdots,\mathbf{z}_{wh}^j] \in \mathbb{R}^{D\times{(wh+1)}}.
\end{equation}
$\{\mathbf{Z}^j\}_{j=1}^L$ are the inputs of our multi-view vision Transformer.

\vspace{-0.1in}
\subsection{Multi-view Vision Transformer}

The proposed multi-view vision Transformer consists of two parts. The first part 
processes patches in each view, individually. It generates the low-level features for patches in each view. We term the first part as local Transformer layers. The second part takes the low-level patch features from the first part as input. It merges patches features from all views in a set and feeds the merged set into a stack of Transformer layers for empowering each patch of each view to have a global reception field. We term these layers exploiting global visual content as global Transformer layers.

\vspace{0.1in}

\noindent \textbf{Local Transformer blocks.} They process the patch features from each view, individually.
Let us denote the number of local Transformer blocks as $S$. The feature matrix $\mathbf{Z}^j$ from each view $V^j$ goes through $S$ blocks sequentially.
The input of the $s$-th local Transformer block is denoted by $\mathbf{Z}^j_{s-1}$  and the output by $\mathbf{Z}^j_{s}$. In this case, the input of the first block $\mathbf{Z}_0^j$ is just the patch feature set $\mathbf{Z}^j$ defined in Eq~\eqref{input}. 
Meanwhile, we denote the layers in the $s$-th local Transformer block by $\{\mathrm{MSA}_s^l,\mathrm{MLP}_s^l, \mathrm{LN}_{s,1}^l,\mathrm{LN}_{s,2}^l\}$. For the $s$-th local Transformer block, it conducts the following operation:
\begin{equation}
    \begin{split}
     \mathbf{Z}_s^j    &\gets  \mathbf{Z}^j_{s-1} + \mathrm{MSA}_s^l(\mathrm{LN}_{s,1}^l(\mathbf{Z}^j_{s-1})),~~~ j = 1 \dots L,\\\mathbf{Z}_s^j   &\gets  \mathbf{Z}^j_s + \mathrm{MLP}_{s}^l(\mathrm{LN}_{s,2}^l(\mathbf{Z}_s^j)),~~~ j = 1 \dots L.
    \end{split}
\end{equation}
After $S$ local Transformer blocks, we obtain the output, $ \mathbf{Z}^j_S$ for each view $V^j$. 

\vspace{0.1in}

\noindent \textbf{Global Transformer blocks.} They process the patch features from each view, jointly. At first, the output of the last local Transformer block for all views, $\{\mathbf{Z}^j_S \}_{j=1}^L$  are concatenated into a global matrix:
\begin{equation}
    \mathbf{M} = [\mathbf{Z}^1_S,\cdots,\mathbf{Z}^L_S ] \in \mathbb{R}^{D\times Lwh}.
\end{equation}
We denote the layers in the $t$-th global Transformer block by $\{\mathrm{MSA}_t^g,\mathrm{MLP}_t^g, \mathrm{LN}_{t,1}^g,\mathrm{LN}_{t,2}^g\}$.  The input of the $t$-th local Transformer block  is denoted by $\mathbf{M}_{t-1}$  and the output is denoted  by $\mathbf{M}_{t}$.  For the $t$-the global Transformer block, it conducts the following operation:
\begin{equation}
    \begin{split}
     \mathbf{M}_t    &\gets  \mathbf{M}_{t-1} + \mathrm{MSA}_t^g(\mathrm{LN}^g_{t,1}(\mathbf{M}_{t-1})),\\\mathbf{M}_t   &\gets  \mathbf{M}_t + \mathrm{MLP}_{t}^g(\mathrm{LN}_{t,2}^g(\mathbf{M}_t)).
    \end{split}
\end{equation}
After $T$ global Transformer blocks, we obtain the output, $ \mathbf{M}_{T}$:
\begin{equation}
    \mathbf{M}_{T} = [\mathbf{m}_0^1,\mathbf{m}_1^1,\cdots,\mathbf{m}_{wh}^1,\mathbf{m}_0^2,\mathbf{m}_1^2,\cdots,\mathbf{m}_{wh}^2,\cdots, \mathbf{m}_0^L,\mathbf{m}_1^L,\cdots,\mathbf{m}_{wh}^L] \in \mathbb{R}^{D\times L(wh+1)},
\end{equation}
where $\mathbf{m}_0^{j}$ denotes the attended  special token for the view $V^j$. We conduct sum-pooling on $\{\mathbf{m}_0^{j}\}_{j=1}^L$ and obtain the global representation for the 3D object:
\begin{equation}
    \mathbf{m} = \frac{1}{L} \sum_{j=1}^L \mathbf{m}_0^{j}.
\end{equation}
Then $\mathbf{m}$ is fed into a fully-connected layer for classification.

\vspace{-0.1in}
\section{Experiments}

\vspace{-0.1in}

\textbf{Datasets.}
We perform experiments on ModelNet40 and ModelNet10~\cite{wu20153d}. ModelNet40 consists of $12,311$ 3D CAD models from 40 categories. 
Among them, $9,843$ models are for training, and $2,468$ models are for testing. ModelNet10 is a subset of ModelNet40 and has 10 categories.
We use two settings for generating views from the 3D object.
The 12-view setting follows the setup in~\cite{wang2017dominant} and the 20-view setting follows the manner in~\cite{kanezaki2018rotationnet}.

\vspace{0.1in}
\noindent \textbf{Implementation.}
The architecture of our MVT follows  DeiT~\cite{touvron2020training}. We also attempt several configurations, including the tiny and small models. The details of different configurations are summarized in Table~\ref{tab:details}.

\begin{table}[h!]
\centering
\begin{tabular}{cccccc}
\toprule
      & hidden dimension & \# heads & \# local blocks & \# global blocks &  \\ \hline
tiny  &      $192$            & $3$         &    $8$             &      $4$            &  \\ 
small &     $384$             &  $6$        &    $8$             &      $4$            &  \\  \toprule
\end{tabular}
\caption{The details of different settings.}
\label{tab:details}\vspace{-0.1in}
\end{table}

\noindent \textbf{Training details.}
The training follows the settings in DeiT~\cite{touvron2020training}. To be specific, we use AadmW~\cite{adamw} as the optimizer, with an initial learning rate of 0.001, $\beta_1$=0.9, $\beta_2$=0.98. Our model is implemented based on the PaddlePaddle deep learning platform. The model is trained with mixed precision on 2 NVIDIA V100 GPUs.  We train 300 epochs for training from scratch and 100 epochs for finetuning on a pre-trained model. The pre-trained models are trained on ImageNet1K dataset~\cite{deng2009imagenet}.

\subsection{The influence of the global Transformer blocks}
When the number of global Transformer blocks is $0$, it is equivalent to replacing the CNN in MVCNN~\cite{su2015multi} by a vision Transformer. We term this configuration as the local baseline. In this baseline setting, each patch can only communicate with patches from the same view. When we replace the local transformer block with the global transformer block, the patch embeddings from the local transformer are fed into the global transformer to interrelate with patches from other views. 

\begin{table}[htp]
\centering
\resizebox{1\linewidth}{!}{
\begin{tabular}{@{}ccccccccccccc@{}}
\toprule
& \multicolumn{6}{c}{tiny} & \multicolumn{6}{c}{small}\\
\cmidrule(lr){2-7} \cmidrule(lr){8-13}
local blocks & 12 & 11 & 10 & 8 & 4 & 0 & 12 & 11 & 10 & 8 & 4 & 0 \\
global blocks & 0 & 1 & 2 & 4 & 8 & 12 & 0 & 1 & 2 & 4 & 8 & 12 \\
\midrule
w/o pre-train & 90.42 & 91.02 & 91.30 & 92.35 & 92.07 & 91.13 & 92.57 & 92.35 & 92.73 & 93.12 & 92.79 & 92.35 \\
w/ pre-train & 94.55 & 94.16 & 94.38 & 94.82 & 94.82 & 94.66 & 94.77 & 94.71 & 95.10 & 95.21 & 95.04 & 95.21 \\
\midrule
GPU memory (M) & 1523 & 1528 & 2114 & 2548 & 3394 & 4289 & 3072 & 3817 & 4242 & 5095 & 6843 & 8586 \\ 
time/epoch (s) & 15 & 17 & 17 & 20 & 23 & 26 & 24 & 25 & 27 & 33 & 40 & 49 \\
throughput (obj/s) & 53 & 50 & 47 & 47 & 36 & 18 & 25 & 23 & 23 & 19 & 16 & 15 \\
\bottomrule
\end{tabular}
}
\vspace{0.0in}

\caption{Evaluation of our method with different numbers of block layers on ModelNet10 dataset, where w/o pre-train denotes training from scratch and w/ pre-train means fine-tuning from a pre-trained model. The view number is fixed as 6. We set the batch size (the number of 3D objects per batch) as 8 when testing the GPU memory cost. The inference throughput is measured as the number of 3D objects processed per second on an NVIDIA TITAN X Pascal GPU. Setting 4 global transformer blocks strikes a good balance between interrelate intra-view and inter-view on patches.
}
\label{tab:block}
\end{table}

\vspace{0.1in}
\noindent \textbf{Accuracy.} To investigate how the number of global transformer blocks and local transformer blocks influence the performance, we use the ModelNet10 as a testbed to ablate. The results are shown in Table~\ref{tab:block}. When training the tiny model from scratch, we can see that when we add global transformer blocks, the performance improves compared to the local baseline, no matter how deep. Training the small model without fine-tuning on the pre-trained model, including global transformer blocks, has better performance in most cases. The increase proves that adding global Transformer blocks to communicate patches from different views helps the classification performance. The local baseline achieves 90.42\% using the tiny model and 93.12\% using the small model. When including 4 global Transformer blocks with the tiny model, the performance achieves the highest with an accuracy of 92.35\%  compared to 90.42\% of local baseline,  compared to 92.57\% of local baseline. However, increasing the number of global Transformer blocks does not always guarantee better performance. When we have more than 4 global Transformer blocks, the accuracy declines. Less local Transformer blocks, leading to fewer layers where patches only attend to intra-view patches, could cause such a decrease. We suppose intra-view attention for low layers is essential. If we alter all local transformer blocks with global transformer blocks, there is no layer to restrict the patches interrelate with other patches only within its view. Using all global transformer blocks is another setting we call global baseline. 
However, when fine-tuned on pre-trained models, more global Transformer blocks do not always lead to better performance.

To investigate the choice of the number of global Transformer blocks, we keep fixed local Transformer blocks while gradually increasing the global Transformer blocks from 0 to 6. Table~\ref{tab:fixed} displays the accuracy on the ModelNet10 test set with a different number of global Transformer blocks. We observe that more global Transformer blocks do not always lead to better performance. The accuracy of 5 and 6 global Transformer blocks is lower than of 4. More global Transformer blocks have more trainable parameters.
From Table~\ref{tab:block} and Table~\ref{tab:fixed} we conclude that 4 global transformer blocks is a good trade-off and we set as the default.

\begin{table}[htp]
\centering
\begin{tabular}{lccccccc}
\toprule
global & 0 & 1 & 2 & 3 & 4 & 5 & 6 \\
\midrule
accuracy & 90.42 & 91.46 & 91.57 & 92.02 & \textbf{92.35} & 91.85 & 91.96 \\
\bottomrule
\end{tabular}

\vspace{0.1in}
\caption{
    Evaluating the number of global Transformer blocks when fixing the number of the local Transformer blocks.
}
\label{tab:fixed}
\end{table}

\noindent \textbf{Efficiency.} Meanwhile, in the last rows of Table~\ref{tab:block}, we show the GPU memory cost per batch and the time cost per epoch of different settings. We include the inference speed in Table~\ref{tab:block} as well. It is shown that when we use more global blocks, the GPU memory and the training time increase accordingly. At the same time, the inference speed decrease correspondingly. Specifically, using the small model, when the number of global blocks increases from $0$ to $12$, the GPU memory cost increases from $3072$M to $8586$M, the training time cost per epoch increases from $24$ seconds to $49$ seconds, and the throughput decreases from $25$ objects/second to $15$ objects/second.
Considering both effectiveness and efficiency, we set the number of the local Transformer blocks as $8$ and the number of the global Transformer block as $4$, by default.

\vspace{0.1in}
\noindent \textbf{Tiny vs Small.} From Table~\ref{tab:block} we can also see the performance improvement of using a larger model. The small model has 6 heads with a hidden dimension of 384. The number of heads and dimensions is double to the tiny model, leading to better performance. Take 4 global Transformer blocks as an example. The accuracy increases from 92.35\% (tiny without pre-train) to 93.12\% (small without pre-train).

\subsection{The influence of the number of projected views}
\textbf{Accuracy.} We evaluate the effect of the number of views on the average instance accuracy of our method on the ModelNet10 dataset.
As shown in Table~\ref{tab:M10view}, more projected views lead to better classification accuracy. 
Specifically, using a single view, a small model with pre-training only achieves a $89.98\%$ recognition accuracy, whereas it achieves a $95.26\%$ recognition accuracy using 12 views. 
It is expected since more views will provide more visual information for a 3D object and benefit the 3D object recognition.

\vspace{0.1in}
\noindent \textbf{Efficiency.} We report the number of views on the GPU memory consumption and time cost per epoch in the last two rows of Table~\ref{tab:M10view}. As shown in the table,  using more views leads to  more computational cost and GPU memory cost. In detail, using the small model, the GPU memory per batch increases from $1010$M to $10700$M when the number of views increases from 1 to 12. We recommend to use only $3$ projected views when the computing resources are limited since it has achieved a excellent accuracy. Meanwhile, we suggest to use $12$ projected  views when computing resources are abundant.

\begin{table}[t]
\centering
\begin{tabular}{ccccccccc}
\toprule
& \multicolumn{4}{c}{tiny} & \multicolumn{4}{c}{small}\\
\cmidrule(lr){2-5} \cmidrule(lr){6-9}
views & 1 & 3 & 6 & 12 & 1 & 3 & 6 & 12\\
\midrule
w/o pre-train & 82.54 & 89.70 & 92.35 & 92.47 & 85.19 & 92.35 & 93.12 & 93.13 \\
w/ pre-train & 90.64 & 94.00 & 94.82 & 95.01 & 89.98 & 94.82 & 95.12 & 95.26  \\
\midrule
GPU memory (M) & 365 & 1457 & 2548 & 5455 & 1010 & 2354 & 5095 & 10700 \\
time/epoch (s) & 5 & 11 & 20 & 33 & 25 & 14 & 33 & 68 \\
\bottomrule
\end{tabular}

\vspace{0.05in}

\caption{Evaluation of our method with different view numbers on ModelNet10 datasets. }
\label{tab:M10view}
\end{table}

\subsection{The influence of the class token}
In the current settings, we feed the attended class token feature in the output of the last Transformer block to the classifier to obtain the recognition result. A choice is to average-pool all attended patch features in the output of the last Transformer block to generate a global representation for classification. We investigate the effectiveness of leveraging class tokens compared to average-pooling patch features. The tiny model is trained from scratch on ModelNet10 with 6 views. In Table~\ref{class_token}, we show the experimental results. The improved performance shows that the attended class token achieves a better performance than its counterpart using the global feature obtained from average-pooling the attended patch features.

\begin{table}[htp!]
\centering
\begin{tabular}{ccc}
\toprule
 & avg\_pool & class\_token \\
\midrule
accuracy & 91.45\% & 92.35\% \\
\bottomrule
\end{tabular}

\vspace{0.1in}

\caption{Comparisons between the global feature from average pooling the attended patch features and  the attend class token feature.}
\label{class_token}
\end{table}

\begin{table}[t!]
\centering
\begin{tabular}{ccccc}
\toprule
Method & Views & ModelNet40 & ModelNet10 \\
\midrule
\multicolumn{4}{c}{Volume-based methods}\\
\midrule
3DShapeNets~\cite{wu20153d} & - & 77.0 & 83.5 \\
VoxNet~\cite{maturana2015voxnet} & - & 83.0  & 92.0\\
Volumetric CNN~\cite{qi2016volumetric} & - & 89.9 &-\\
3D-A-Nets~\cite{ren20173d} & - & 90.5 & - \\
LP-3DCNN~\cite{LP-3DCNN} & - & 92.1 & - \\
\midrule
\multicolumn{4}{c}{Point-based methods}\\
\midrule
PointNet~\cite{pointnet}  & - & 89.2 & - \\
PointNet++~\cite{pointnet++}  & - & 91.9 & - \\
3DmFV-Net~\cite{3Dmfv}  & - & 91.6 & 95.2 \\
DeepCCFV~\cite{AAAI2019DeepCCFV} & - & 92.5 & - \\
\midrule
\multicolumn{4}{c}{View-based methods}\\
\midrule
MVCNN~\cite{su2015multi} & 80 & 90.1  & - \\
RotationNet~\cite{kanezaki2018rotationnet} & 12 & 91.0 & 94.0 \\
RotationNet~\cite{kanezaki2018rotationnet}& 20 & 97.4 & 98.5 \\
Relation Network~\cite{mvrelations} & 12 & 94.3 & 95.3 \\
3D2SeqViews~\cite{DBLP:journals/tip/HanLLVLZHC19} & 12 & 93.4 & 94.7 \\
SeqViews2SeqLabels~\cite{DBLP:journals/tip/HanSLVLZHC19} & 12 & 93.4 & 94.8 \\
GVCNN~\cite{feng2018gvcnn} & 12 & 93.1 & - \\
CAR\-Net~\cite{DBLP:journals/tip/XuZXQL21} & 12 & 95.2 & 95.8 \\
CAR\-Net~\cite{DBLP:journals/tip/XuZXQL21} & 20 & \textbf{97.7} & 99.0 \\
{\textbf{MVT-small (Ours)}} & 12 & 94.4 & 95.3\\
{\textbf{MVT-small (Ours)}} & 20 & 97.5  & \textbf{99.3} \\
\bottomrule
\end{tabular}

\vspace{0.1in}

\caption{
Comparison with the present state-of-the-art methods on ModelNet40 dataset.
}
\label{tab:sota}
\end{table}

\subsection{Comparsions with state-of-the-art methods}
We compare with three groups of methods including volume-based methods, point-based methods and view-based methods in Table~\ref{tab:sota}. The first part of  Table~\ref{tab:sota} reports the performance of volume-based methods including 3DShapeNets~\cite{wu20153d}, VoxNet~\cite{maturana2015voxnet}, Volumetric CNN~\cite{qi2016volumetric}, 3D-A-Nets~\cite{ren20173d}, and LP-3DCNN~\cite{LP-3DCNN}. As shown in the table, the recognition accuracy of these volume-based methods are not competitive compared with view-based methods.

Then we compare with point-based methods including PointNet~\cite{pointnet}, PointNet++~\cite{pointnet++}, 3DmFV-Net~\cite{3Dmfv}, and DeepCCFV~\cite{AAAI2019DeepCCFV}. Compared with volume-based methods, point-based methods achieve  considerably higher recognition accuracy. To be specific, PointNet++~\cite{pointnet++} achieves $91.9$ recognition accuracy. It significantly outperforms the best volume-based method in Table~\ref{tab:sota}, 3D-A-Nets~\cite{ren20173d},  with only $90.5$ recognition accuracy. But the point-based methods are still not as competitive as their view-based counterparts.

At last, we compare with view-based methods including MVCNN~\cite{su2015multi}, RotationNet~\cite{kanezaki2018rotationnet}, 3D2SeqViews~\cite{DBLP:journals/tip/HanLLVLZHC19}, SeqViews2SeqLabels~\cite{DBLP:journals/tip/HanSLVLZHC19}, Relation Network~\cite{mvrelations} and CAR\-Net~\cite{DBLP:journals/tip/XuZXQL21} on both 12-view and 20-view settings. 
With more views, the performance achieved using  $20$-view settings usually is better than $12$-view settings. Compared with these methods, our MVT-small model achieves competitive performance. Specifically, on the ModelNet10 dataset, using 20-view settings, we reach the highest recognition accuracy. It is worth noting that our MVT-small architecture is conceptually simple with more minor hand-designed components than the compared methods such as Relation Network~\cite{mvrelations} and CAR\-Net~\cite{DBLP:journals/tip/XuZXQL21}.  

\section{Conclusion}
In this paper, we propose a multi-view vision Transformer (MVT) for effective 3D object recognition. Considering the efficiency, we design our MVT in a local-global structure. The global Transformer layers empower each patch to communicate with the patches from all views, overcoming the limitations of existing CNN-based models with a local reception field on patches from the same view. Although the proposed MVT is in a conceptually simple structure, it has achieved state-of-the-art recognition performance on public benchmarks, including ModelNet40 and ModelNet10 datasets.

\bibliography{custom}
\end{document}